%% file: CameraReady.tex
\documentclass[11pt]{article}

\usepackage[final]{acl}

\usepackage{times}
\usepackage{latexsym}

\usepackage[T1]{fontenc}

\usepackage[utf8]{inputenc}

\usepackage{microtype}

\usepackage{inconsolata}

\usepackage{graphicx}
\usepackage{enumitem}

\input{latex/preamble}
\title{
Learning from Another Angle: Rotation-Based Test-Time Training for Multimodal Large Language Models
}

\author{
 \textbf{Changyu Liu\textsuperscript{1\thanks{Equal contribution}}},
 \textbf{Fei Zhao\textsuperscript{2\footnotemark[1]}},
  \textbf{Yiyang Liu\textsuperscript{1}},
 \textbf{Xi Xiao\textsuperscript{2}},
 \textbf{Tianyang Wang\textsuperscript{2\footnotemark[2]}},
 \textbf{Cheng Han\textsuperscript{1\thanks{Corresponding authors}}}
\\
\\
 \textsuperscript{1}University of Missouri--Kansas City,
 \textsuperscript{2}University of Alabama at Birmingham\\
}

\begin{document}
\maketitle
\input{sec/0_abstract}
\input{sec/1_intro}
\input{sec/2_rl}

\input{sec/3_method}
\input{sec/4_exp}

\input{sec/6_conclusion}

\input{sec/limitation}
\input{sec/acknowledge}

\bibliography{latex/custom}

\input{sec/X_suppl}

\end{document}

%% file: latex/preamble.tex
\usepackage{xcolor}

\definecolor{myred}{RGB}{255 0 0} 
\definecolor{mygreen2}{RGB}{0 176 80} 
\definecolor{mygreen3}{RGB}{80,99,43} 
\definecolor{mygray}{gray}{.9}

\makeatletter
\newcommand{\thickhline}{%
    \noalign {\ifnum 0=`}\fi \hrule height 1pt
    \futurelet \reserved@a \@xhline
}

\usepackage{adjustbox}
\usepackage{pifont}

\usepackage{multirow}
\usepackage{xcolor}
\usepackage{colortbl}
\usepackage{booktabs}
\usepackage{algorithm}
\usepackage{algorithmic}
\usepackage{listings}

\usepackage{amsthm}

\definecolor{mygray}{gray}{.9}

\newcolumntype{Y}{>{\raggedleft\arraybackslash}X}

\usepackage{times}    
\usepackage{xspace}
\usepackage{graphicx}
\usepackage{amsmath}
\usepackage{amssymb}
\usepackage{scalerel}
\usepackage{tabularx}

\definecolor{ForestGreen}{RGB}{27,94,32}

\DeclareRobustCommand\onedot{\futurelet\@let@token\@onedot}
\def\@onedot{\ifx\@let@token.\else.\null\fi\xspace}
\def\eg{\emph{e.g}\onedot} 
\def\ie{\emph{i.e}\onedot} 

%% file: sec/0_abstract.tex
\begin{abstract}
Multimodal Large Language Models (MLLMs) achieve strong performance on vision-language tasks but often remain vulnerable to distribution shifts due to their fixed parameters during inference. 
While Test-Time Training (TTT) offers a pathway for dynamic adaptation, existing approaches suffer from prohibitive computational overhead due to multiple sampling and the risk of error reinforcement from biased self-generated supervision.
In light of this view, we propose RoTTT (\textbf{Ro}tation-based \textbf{T}est-\textbf{T}ime \textbf{T}raining), a simple and efficient TTT approach that enables instance-specific refinement at inference time. For each test instance, RoTTT performs a small number of gradient updates by predicting the rotation angle of the input image.
The known rotation angles naturally serve as a reliable self-supervised signal, enabling lightweight and stable adaptation without requiring additional annotations.
Extensive experiments across multiple vision-language benchmarks demonstrate that RoTTT consistently improves performance across different MLLMs, providing a practical and efficient paradigm for on-the-fly MLLM adaptation.
\end{abstract}

%% file: sec/1_intro.tex
\vspace{-2mm}
\section{Introduction}
\label{sec:intro}
Multimodal Large Language Models (MLLMs) have demonstrated strong performance across a wide range of vision-language tasks, including visual question answering (VQA), captioning, and reasoning. However, a common observation is that their performance often degrades under distribution shift, where test-time inputs differ from the data seen during pretraining or instruction tuning. Such shifts may arise from variations in visual domains (\eg, lighting, viewpoint, object layout), compositional complexity, or task-specific semantics. A key reason for this degradation is that pretrained MLLMs are typically deployed with fixed parameters, preventing them from adapting to distribution shifts during inference. Similar challenges have been observed in vision and language domains, where recent work~\cite{zuo2025ttrl,hu2025test, wang2020tent, sun2020test} has explored Test-Time Training (TTT) solutions to update model parameters on-the-fly using test data, often without requiring ground-truth labels.

\begin{figure}[t]
    \centering
    \includegraphics[width=0.43\textwidth]{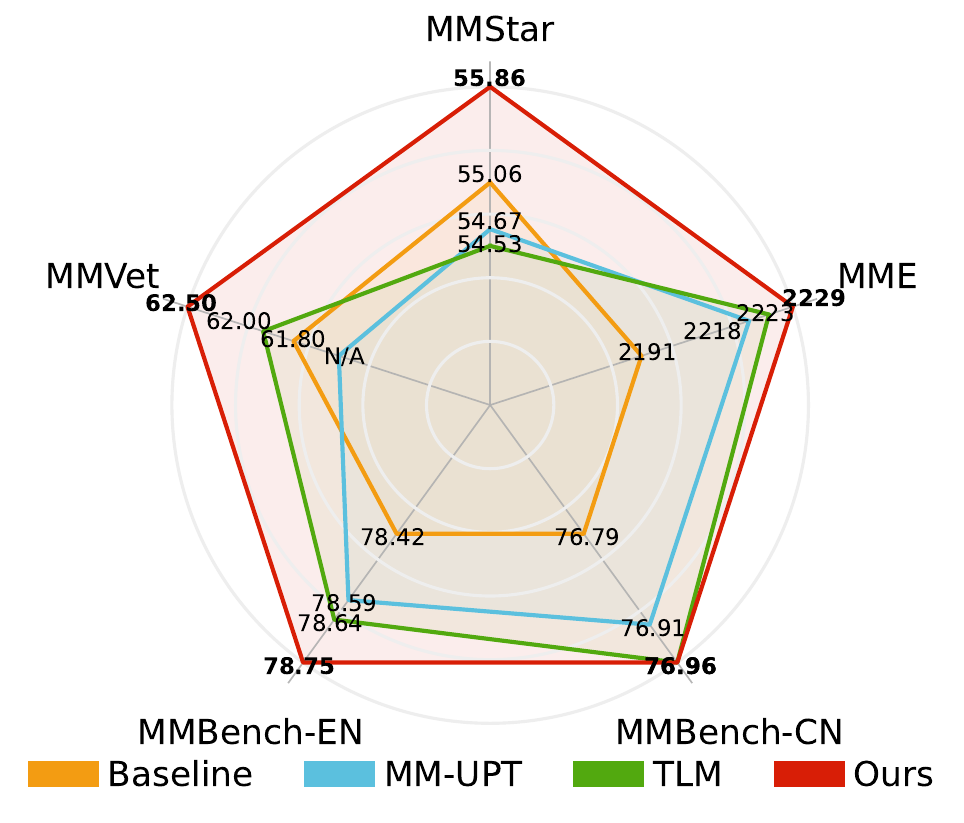}
    \vspace{-6pt}
    \caption{\textbf{Comparison of RoTTT with other test-time training methods} (\ie, MM-UPT~\cite{wei2025first} and TLM~\cite{hu2025test}) on Qwen2.5VL-3B~\cite{Bai2025Qwen25VLTR}. As seen, RoTTT consistently achieves the best results across benchmarks, demonstrating superior effectiveness and stability. More visualizations are provided in Appendix \S\ref{appendix:vis_result}.}
    \vspace{-20pt}
    \label{fig:radar_illustration}
\end{figure}

Despite its great potential, TTT for MLLMs remains largely underexplored. An early post-training attempt~\cite{wei2025first} that directly borrows from GRPO-style optimization and, despite providing performance gains, suffers from \textit{two fundamental limitations}. 
\textit{First}, it relies on multiple sampling, requiring the model to generate and evaluate multiple candidate outputs per instance, resulting in substantial computational overhead. 
\textit{Second,} it updates the model using its own generated predictions as supervision. Under distribution shift, such predictions are generally biased or incorrect, and optimizing over them can reinforce and amplify existing errors rather than correct them. 
These facts highlight the need for a lightweight, reliable alternative for MLLM Test-Time Training.

Our solution is inspired by an interesting human cognitive observation: when humans get stuck on a challenging visual problem, they often change their viewing perspective (\eg, by rotating an object or reorienting their viewpoint) to gain new insights. Crucially, such transformations preserve the underlying semantic content while altering the spatial organization of the scene, potentially revealing previously overlooked structural cues. This suggests that controlled viewpoint changes can induce meaningful representation restructuring without introducing additional semantic ambiguity. Moreover, these transformations naturally provide supervision, as the applied transformation parameters (\eg, rotation angles) are known by construction, enabling sample-specific self-supervision at test time.

In light of this view, we propose RoTTT (\textbf{Ro}tation-based \textbf{T}est-\textbf{T}ime \textbf{T}raining), a simple and instance-aware TTT strategy based on rotation prediction. For each test image, we generate a rotated view and use the known rotation angle as an auxiliary supervision signal. We then conduct a small number of gradient updates on the current instance and use the adapted model to perform the original task, as a practical analogy to human cognition via controlled viewpoint changes. This approach is effective for two reasons. \textit{\textbf{First,}} predicting rotation angles promotes spatial structure awareness, as it requires the model to attend to cues such as object orientation, layout, boundaries, and relative positioning. \textit{\textbf{Second,}} the adaptation is instance-specific, enabling targeted refinement on the current sample. In effect, the model is encouraged to ``re-examine'' the image, leading to improved alignment and more accurate predictions under distribution shift. 

The primary contributions of this work are threefold. First, we propose an efficient TTT paradigm for MLLMs that enables on-the-fly model refinement via rotation-based optimization, without requiring costly sampling or external supervision. Second, we show that rotation prediction provides a free, sample-specific, and spatially-aware self-supervised signal. Third, through extensive experiments, 
we demonstrate that RoTTT consistently improves performance (\eg, up to 2.00 on MMVet), highlighting its robustness and generalizability.

%% file: sec/2_rl.tex
\section{Related Work}
\label{sec:rl}

\subsection{Test-time Training}
Test-time training (TTT)~\cite{sun2020test,hu2025test,yoonc,xiao2025dynaprompt,zhu2024efficient,zuo2025ttrl,liu2026fly} has recently emerged as a powerful paradigm for improving model robustness against distribution shifts. In contrast to conventional training paradigms that rely solely on fixed pretrained models, TTT enables instance-specific adaptation, allowing models to dynamically adjust to unseen inputs without access to source data. While the efficacy of TTT has been rigorously validated in computer vision~\cite{wang2020tent, sun2020test,wang2022continual,xiao2025dynaprompt,wu2024test,iwasawa2021test,zhang2022memo, lim2023ttn, bartler2022mt3,gandelsman2022test, sinha2023test} and large language models (LLMs)~\cite{hu2025test,yoonc,zuo2025ttrl,li2025implicit, hardt2024test, hubotter2025efficiently, behrouz2026titans,wang2024greater,sun2024learning} domains, its application to MLLMs remains a largely underexplored frontier.

Initial attempts adopt reinforcement learning frameworks (\ie, GRPO)~\cite{wei2025first}, which update the model using relative preference signals over multiple sampled responses. However, these efforts suffer from two key limitations: \textit{\textbf{\ding{172}}} significant computational overhead introduced by iterative sampling (see \S\ref{sec:ab_study}), and \textit{\textbf{\ding{173}}} a reliance on self-generated supervision that is inherently prone to error reinforcement~\cite{guo2025deepseek,bai2025m, zhang2025vl}. To circumvent these limitations, we propose a simple and efficient test-time training method for MLLMs that avoids costly multi-sample optimization.

\subsection{Self-supervised Auxiliary Tasks}
Self-supervised auxiliary tasks provide label-free supervision by deriving training signals directly from the input data, and they have played a central role in representation learning. Early work in vision explores pretext objectives, such as predicting contextual relative positions \cite{doersch2015unsupervised,wang2023position,nezhad2025self}, solving jigsaw puzzles \cite{noroozi2016unsupervised}, rotation prediction \cite{gidaris2018rotation}, and reconstructing masked image patches \cite{he2022mae,han2024mat}, all of which encourage the learning of spatial and semantic representations. In parallel, language models rely on token-level auxiliary objectives, such as masked token prediction \cite{devlin2019bert} and next-token prediction \cite{radford2019language}, to capture contextual and syntactic structure during pretraining. These self-supervised objectives are appealing because they require no additional annotations and provide stable supervision directly from the data. However, most prior auxiliary tasks are designed for pretraining or fine-tuning, and their potential as test-time optimization signals for MLLMs remains largely underexplored. To bridge this gap, we leverage deterministic, rotation-based self-supervision to drive instance-specific test-time training for MLLMs.

%% file: sec/3_method.tex
\section{Method}
\label{sec:method}

\begin{figure*}[t]
    \centering
    \includegraphics[width=\textwidth]{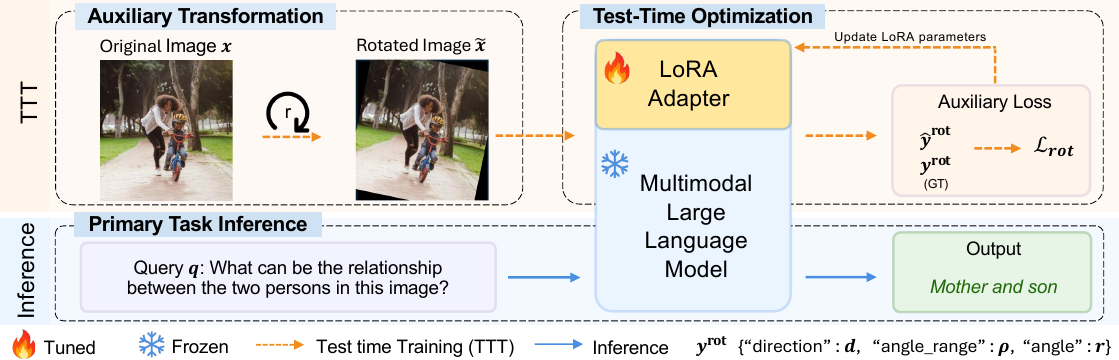}
    \vspace{-15pt}
    \caption{\textbf{Overview of Rotation-based Test-Time Training (RoTTT).} Given a test instance $(x,q)$, RoTTT first generates an auxiliary transformed view $\tilde{x}$ by applying a controlled rotation to the input image $x$. The image pair $(x,\tilde{x})$ is then processed by the pretrained MLLM with trainable LoRA parameters to perform a structured textual rotation prediction task, where the supervision signal $y^{\text{rot}}$ is directly derived from the rotation parameter $r$. The resulting auxiliary loss $\mathcal{L}_{\text{rot}}$ is used to update the LoRA parameters during test-time training. After adaptation, the updated MLLM performs inference on the original image $x$ and query $q$ to generate the final prediction.}
    \vspace{-10pt}
    \label{fig:main_figure}
\end{figure*}

\subsection{Problem Formulation}
Consider a Multimodal Large Language Model (MLLM) $f_\theta$, parameterized by $\theta$, which maps an input image $x \in \mathcal{X}$ and a textual query $q \in \mathcal{Q}$ to a response $y \in \mathcal{Y}$. 
The response, generally generated in an autoregressive way, can be denoted as:
\begin{equation}
P(y \mid x, q; \theta) = \prod_{t=1}^{T} P(y_t \mid x, q, y_{<t}; \theta),
\end{equation}
where $y_t$ is the $t$-th token in the output sequence of length $T$.

For a fixed MLLM during inference, samples $(x, q) \sim \mathcal{D}_{\text{test}}$ may come from a distribution different from the original training distribution $\mathcal{D}_{\text{train}}$, leading to performance degradation under distribution shift~\cite{zhang2024out,verma2024evaluating}. 
TTT aims to adapt the model to $\mathcal{D}_{\text{test}}$ by updating its parameters for each test instance without access to source data or ground-truth labels. Given a test sample $(x, q)$, the model parameters are updated from $\theta$ to $\theta'$ and the updated model $f_{\theta'}$ is then applied for the sample-specific prediction. 

The overall training objective is to minimize the expected task loss $\mathcal{L}_{\text{task}}$ over the test distribution:
\begin{equation}
\mathbb{E}_{(x,q)\sim \mathcal{D}_{\text{test}}} \left[ \mathcal{L}_{\text{task}} \big(f_{\theta'}(x,q), y^* \big) \right],
\end{equation}
where $y^*$ denotes the ground-truth output. Since $y^*$ is unavailable at test time, the task loss $\mathcal{L}_{\text{task}}$ cannot be directly optimized. Instead, TTT introduces a self-supervised auxiliary objective $\mathcal{L}_{\text{aux}}$ as:
\begin{equation}\label{eq:aux}
\theta' = \arg\min_{\theta} \mathcal{L}_{\text{aux}}(x; \theta).
\end{equation}

The optimization is typically performed via a limited number of gradient descent iterations.  
As shown in Eq.~\ref{eq:aux}, the effectiveness of TTT fundamentally depends on the design of the self-supervised auxiliary objective (see \S\ref{sec:ab_study}).

\subsection{Rotation-based Test-Time Training (RoTTT)}
The efficacy of TTT hinges on the reliability of the supervision signal at inference. Existing approaches~\cite{wei2025first} for MLLMs that use GRPO-style optimization typically depend on self-generated outputs to obtain these signals. However, this paradigm faces a dual challenge: 
\textit{\textbf{\ding{172}}} Substantial computational overhead due to repeated sampling, and
\textit{\textbf{\ding{173}}} Unreliable supervision, as updates are driven by the model’s own predictions, which can be biased or incorrect under distribution shift, thereby amplifying errors. These challenges suggest that an effective test-time signal should be label-free, reliable, and computationally efficient. Motivated by the cognitive observation that humans change their viewing perspective to gain new insights into challenging visual problems, we turn to leverage transformation-based self-supervision. Specifically, we apply controlled transformations (\ie, rotation) to each input image and use the known transformation parameters as supervision, formulating test-time training as an auxiliary transformation prediction problem that provides a lightweight and reliable objective for instance-specific refinement.

\noindent\textbf{Rotation-based Auxiliary Task.} Given a test image $x$, we construct an auxiliary self-supervised task by applying a controlled rotation transformation. Specifically, we sample a rotation angle $r$ from a predefined angle space $\mathcal{A}$, and obtain the transformed image:
\begin{equation}
\tilde{x} = \mathcal{R}_r(x), \quad r \in \mathcal{A},
\end{equation}
where $\mathcal{R}_r(\cdot)$ denotes a rotation operation parameterized by angle $r$. In our implementation, the sign of $r$ indicates the rotation direction: positive values correspond to clockwise rotations and negative values to counter-clockwise rotations. This process yields an auxiliary image pair $(x, \tilde{x})$. 
Because the transformation is generated during construction, the rotation angle parameter $r$ provides a deterministic supervision signal, serving as a ``ground-truth by construction'' that requires no external annotations or model-based heuristics. 

\noindent\textbf{Textual Rotation Objective.}
Instead of formulating rotation prediction as a standalone classification problem with an additional task-specific head, we treat it as a structured textual reasoning task. This approach ensures that the auxiliary task remains compatible with the MLLM's native generative interface. Specifically, 
the model is presented with the image pair $(x, \tilde{x})$ and a rotation prediction prompt $p_{\text{rot}}$ directing it to predict the applied rotation.
To facilitate stable and precise optimization, we define a structured output format $y_{rot}$ consisting of three hierarchical fields: (i) the rotation direction, (ii) a coarse angle range, and (iii) the exact rotation angle. Formally, we define the rotation target $y_{rot}$ as a structured sequence of semantic attributes, encoded in a JSON format:
\begin{equation*}
\{\text{``direction''}: d,\ \text{``angle\_range''}: \rho,\ \text{``angle''}: r\}
\end{equation*}
where $d \in \{\text{``clockwise'', ``counter-clockwise''}\}$, $\rho \in \{\text{``small'', ``medium'', ``large''}\}$, and $r \in \mathbb{Z}$ represents the discrete rotation degree. 

This hierarchical formulation decomposes the geometric reasoning task into coarse-to-fine levels of granularity, providing richer supervision than a monolithic label (see \S\ref{sec:ab_study}). Crucially, by operating the auxiliary task within the textual token space, we ensure that the optimization objective is intrinsically aligned with the model's native autoregressive language modeling objective~\cite{liu2023visual}. This avoids the need for heterogeneous loss functions and allows the model to ``re-align'' its internal spatial representations using its existing language modeling capabilities.

\noindent\textbf{Test-Time Optimization.}
Given the auxiliary image pair $(x, \tilde{x})$, the rotation prediction prompt $p_{\text{rot}}$, and the structured ground-truth target sequence $y^{\text{rot}} = (y_1, \dots, y_T)$, we optimize the MLLM using a standard autoregressive cross-entropy objective. The auxiliary loss $\mathcal{L}_{\text{rot}}$ is defined as:
\begin{equation}\small
\mathcal{L}_{\text{rot}}(x, \tilde{x}, y^{\text{rot}})
= - \sum_{t=1}^{T} \log P\big(y^{\text{rot}}_t \mid x, \tilde{x}, p^{\text{rot}}, y^{\text{rot}}_{<t} \big),
\end{equation}
where $y_{<t}^{\text{rot}}$ denotes the prefix of the target rotation sequence. At test time, we perform a small number of gradient-based adaptation steps on the current test instance:
\begin{equation}
\theta^{k+1}
=
\theta^{k}
-
\eta \nabla_{\theta^{k}} \mathcal{L}_{\text{rot}}(x, \tilde{x}, y^{\text{rot}}),
\end{equation}
where $\eta$ denotes the learning rate. We initialize $\theta^{(0)}$ with the pretrained parameters $\theta$, and denote the final adapted parameters as $\theta'$.
The adapted model is then used to answer the original multimodal task on the untransformed input:
\begin{equation}
\hat{y} = f_{\theta'}(x, q).
\end{equation}

\noindent\textbf{Overall Pipeline.} Fig.~\ref{fig:main_figure} illustrates an overview of RoTTT. For each test instance $(x, q)$, we construct a rotated view $\tilde{x} = \mathcal{R}_r(x)$ and derive the corresponding structured target $y^{\text{rot}}$ based on the deterministic rotation parameter $r$. The MLLM is then refined via instance-specific adaptation, where the parameters are updated by minimizing the auxiliary objective $\mathcal{L}_{\text{rot}}$ on the triplet $(x, \tilde{x}, p_{\text{rot}})$ for a limited number of gradient descent steps. This process yields the instance-specific parameters $\theta'$. Finally, the adapted model is deployed to address the original query $q$ within the unmodified input space, yielding the output $\hat{y} = f_{\theta'}(x, q)$. The pseudo code is shown in Algorithm~\ref{alg:tt_rotation}.

Our proposed rotation-based test-time training framework enjoys several appealing properties:
\begin{itemize}
    \item \textbf{\textit{Efficiency and Simplicity:}} The auxiliary task is constructed via a simple rotation transformation, requiring no additional data. In addition, the optimization follows a standard autoregressive objective and introduces no architectural modifications or task-specific modules, enabling seamless integration with existing MLLMs. Moreover, unlike GRPO-style approaches, our method avoids multiple candidate sampling, thereby significantly reducing computational overhead (see \S\ref{sec:ab_study}). 
    \item \textbf{\textit{Reliable Supervision:}} Our method benefits from a reliable supervision signal derived from transformation parameters that are known by construction. The rotation angle provides a deterministic and noise-free target, ensuring that gradient updates are consistently aligned with the underlying transformation. This leads to stable and well-behaved optimization dynamics during test-time training. To further validate this property, in \S\ref{sec:obs}, we further analyze the loss landscape induced by our auxiliary objective.
    \item \textbf{\textit{Spatial Awareness:}} The rotation prediction task inherently requires understanding spatial configuration, including object orientation, layout, and boundaries. This enforces attention to geometric cues and promotes the extraction of spatially-aware features. As a result, the optimization process induces an implicit restructuring of internal representations, improving their alignment with the underlying visual structure. This behavior is consistent with our empirical observations in \S\ref{sec:obs}.
\end{itemize}

\begin{algorithm}[t]
\caption{RoTTT Pipeline}
\label{alg:tt_rotation}
\begin{algorithmic}[1]
    \STATE \textbf{Input:} Pretrained MLLM $f_\theta$, a test instance $(x, q)$, rotation angle space $\mathcal{A}$, auxiliary prompt $p_{\text{rot}}$, learning rate $\eta$, update steps $K$.
    \STATE \textbf{Output:} Predicted response $\hat{y}$.
    
    \STATE \textbf{Step 1: Auxiliary Transformation}
    \STATE Sample a rotation angle $r \in \mathcal{A}$
    \STATE Generate transformed image $\tilde{x} = \mathcal{R}_r(x)$
    \STATE Construct structured ground-truth sequence $y^{\text{rot}}$ based on $r$
    
    \STATE \textbf{Step 2: Test-Time Optimization}
    \STATE Initialize adapted parameters $\theta^{0} \leftarrow \theta$
    \FOR{$k = 0$ \TO $K-1$}
        \STATE Compute auxiliary loss $\mathcal{L}_{\text{rot}}(x, \tilde{x}, y^{\text{rot}})$ using Eq. (5)
        \STATE Update parameters: $\theta^{k+1} \leftarrow \theta^{k} - \eta \nabla_{\theta^{k}} \mathcal{L}_{\text{rot}}$
    \ENDFOR
    \STATE Set $\theta' \leftarrow \theta^{K}$
    
    \STATE \textbf{Step 3: Primary Task Inference}
    \STATE Generate final output $\hat{y} = f_{\theta'}(x, q)$ on original input
    \RETURN $\hat{y}$
\end{algorithmic}
\end{algorithm}

%% file: sec/4_exp.tex
\section{Experiment}
\label{sec:exp}

\begin{table*}[ht]
\begin{adjustbox}{width=0.99\width,center}
\begin{tabular}{l||ccccc} 
\hline \thickhline
\rowcolor{mygray}
Method &  MMVet & MMStar &  MME  & MMbench-CN &  MMbench-EN \\
\hline \hline

$\text{Qwen2.5VL-3B}$\textcolor{lightgray}{\scriptsize{[arXiv25]}}   &61.80  &55.06  &2191  &76.79  &78.42\\
 + MM-UPT \textcolor{lightgray}{\scriptsize{[NeurIPS25]}} &-  &54.67\small{\textcolor{myred}{$\downarrow$0.39}}  &2218\small{\textcolor{ForestGreen}{$\uparrow$27}}  &76.91\small{\textcolor{ForestGreen}{$\uparrow$0.12}}  &78.59\small{\textcolor{ForestGreen}{$\uparrow$0.17}}\\
 + TLM \textcolor{lightgray}{\scriptsize{[ICML25]}} &62.00\small{\textcolor{ForestGreen}{$\uparrow$0.20}}  &54.53\small{\textcolor{myred}{$\downarrow$0.53}}  &2223\small{\textcolor{ForestGreen}{$\uparrow$32}}  &\textbf{76.96\small{\textcolor{ForestGreen}{$\uparrow$0.17}}} &78.64\small{\textcolor{ForestGreen}{$\uparrow$0.22}}\\
  \rowcolor{mygray}
 + Ours &\textbf{\textbf{62.50\small{\textcolor{ForestGreen}{$\uparrow$0.70}}}}  &\textbf{55.86\small{\textcolor{ForestGreen}{$\uparrow$0.80}}}  &\textbf{2229\small{\textcolor{ForestGreen}{$\uparrow$38}}}  &\textbf{76.96\small{\textcolor{ForestGreen}{$\uparrow$0.17}}}  &\textbf{78.75\small{\textcolor{ForestGreen}{$\uparrow$0.33}}}\\
 \hline
 $\text{Qwen2.5VL-7B}$\textcolor{lightgray}{\scriptsize{[arXiv25]}}   &67.30  &63.93 & 2327 &81.33 &82.06\\
 + MM-UPT \textcolor{lightgray}{\scriptsize{[NeurIPS25]}} &-  &63.93\small{\textcolor{ForestGreen}{$\uparrow$0.00}}    &2321\small{\textcolor{myred}{$\downarrow$6}}  &81.33\small{\textcolor{ForestGreen}{$\uparrow$0.00}}  &82.17\small{\textcolor{ForestGreen}{$\uparrow$0.11}}\\
 + TLM \textcolor{lightgray}{\scriptsize{[ICML25]}} &67.80\small{\textcolor{ForestGreen}{$\uparrow$0.50}}  &63.93\small{\textcolor{ForestGreen}{$\uparrow$0.00}}  &2322\small{\textcolor{myred}{$\downarrow$5}}  &81.45\small{\textcolor{ForestGreen}{$\uparrow$0.12}}  &82.17\small{\textcolor{ForestGreen}{$\uparrow$0.11}}\\
  \rowcolor{mygray}
 + Ours &\textbf{67.90\small{\textcolor{ForestGreen}{$\uparrow$0.60}}}  &\textbf{64.33\small{\textcolor{ForestGreen}{$\uparrow$0.40}}}  &\textbf{2329\small{\textcolor{ForestGreen}{$\uparrow$2}}}  &\textbf{81.50\small{\textcolor{ForestGreen}{$\uparrow$0.17}}}  &\textbf{82.23\small{\textcolor{ForestGreen}{$\uparrow$0.17}}}\\
 \hline
 $\text{LLaVA-NeXT-7B}$ \textcolor{lightgray}{\scriptsize{[arXiv24]}}   &42.60  &37.53  &1788  &60.99  &65.69\\
 + MM-UPT \textcolor{lightgray}{\scriptsize{[NeurIPS25]}} &-  &37.00\small{\textcolor{myred}{$\downarrow$0.53}} &1766\small{\textcolor{myred}{$\downarrow$22}}  &\textbf{61.35\small{\textcolor{ForestGreen}{$\uparrow$0.36}}}  &65.69\small{\textcolor{ForestGreen}{$\uparrow$0.00}}\\
 + TLM \textcolor{lightgray}{\scriptsize{[ICML25]}} &43.90\small{\textcolor{ForestGreen}{$\uparrow$1.30}}   &37.73\small{\textcolor{ForestGreen}{$\uparrow$0.20}}  &1787\small{\textcolor{myred}{$\downarrow$1}}  &60.99\small{\textcolor{ForestGreen}{$\uparrow$0.00}}  &65.75\small{\textcolor{ForestGreen}{$\uparrow$0.06}}\\
  \rowcolor{mygray}
 + Ours &\textbf{44.60\small{\textcolor{ForestGreen}{$\uparrow$2.00}}}  &\textbf{38.00\small{\textcolor{ForestGreen}{$\uparrow$0.47}}}  &\textbf{1811\small{\textcolor{ForestGreen}{$\uparrow$23}}}  &61.27\small{\textcolor{ForestGreen}{$\uparrow$0.28}}  &\textbf{65.92\small{\textcolor{ForestGreen}{$\uparrow$0.23}}}\\
 \hline

 $\text{LLaVA-NeXT-13B}$ \textcolor{lightgray}{\scriptsize{[arXiv24]}}   &45.20  &39.67  &1664  &60.54  &67.43\\
 + MM-UPT \textcolor{lightgray}{\scriptsize{[NeurIPS25]}} &-  &- &- &-  &-\\
 + TLM \textcolor{lightgray}{\scriptsize{[ICML25]}}
 &44.60\small{\textcolor{myred}{$\downarrow$0.60}} &39.60\small{\textcolor{myred}{$\downarrow$0.07}}  &1690\textbf{\small{\textcolor{ForestGreen}{$\uparrow$26}}}  &60.48\small{\textcolor{myred}{$\downarrow$0.06}} &67.49\textbf{\small{\textcolor{ForestGreen}{$\uparrow$0.06}}} \\
  \rowcolor{mygray}
 + Ours &\textbf{45.30\textbf{\small{\textcolor{ForestGreen}{$\uparrow$0.10}}}}  &\textbf{39.87\textbf{\small{\textcolor{ForestGreen}{$\uparrow$0.20}}}}  &\textbf{1693\textbf{\small{\textcolor{ForestGreen}{$\uparrow$29}}}}  &\textbf{60.77\textbf{\small{\textcolor{ForestGreen}{$\uparrow$0.23}}}}  &\textbf{67.54\textbf{\small{\textcolor{ForestGreen}{$\uparrow$0.11}}}}\\
 \hline

\end{tabular}
\end{adjustbox}

\caption{\textbf{Main results on multimodal understanding benchmarks.} We report performance across five benchmarks, including MMVet, MMStar, MME, MMBench-CN, and MMBench-EN, evaluated on four representative MLLMs (\ie, Qwen2.5VL-3B, and Qwen2.5VL-7B, LLaVA-NeXT-7B, and LLaVA-NeXT-13B,). \textcolor{ForestGreen}{$\uparrow$} and \textcolor{myred}{$\downarrow$} denote absolute improvements and declines with respect to the corresponding base model, respectively. Best results are highlighted in \textbf{bold}. The symbol ``-'' represents missing results under specific configurations: MM-UPT is inherently incompatible with the open-ended generative format of MMVet, and it suffers from out-of-memory (OOM) failures on LLaVA-NeXT-13B due to its GRPO-based optimization. As shown, our method consistently improves performance across all models and benchmarks, demonstrating strong effectiveness, stability, and generalization.}
\vspace{-3mm}
\label{table:main_result}
\end{table*}
We conduct extensive experiments to evaluate the effectiveness of our method across multiple MLLMs and diverse multimodal understanding benchmarks. We first introduce the experimental setup, including datasets, baselines, and implementation details (\S\ref{sec:exp setup}). We then present the main results (\S\ref{sec:main_result}), empirical observations (\S\ref{sec:obs}), and ablation studies (\S\ref{sec:ab_study}) to provide a comprehensive analysis of our approach.
More experimental details are provided in Appendix \S\ref{appendix:datasets}-\ref{appendix:add_comp_cost_result}.
\subsection{Experimental Setup}\label{sec:exp setup}
\noindent \textbf{Datasets.} 
We evaluate our method on five prominent multimodal understanding benchmarks: MMVet~\cite{yu2024mm}, MME~\cite{fu2023mme}, MMStar~\cite{chen2024we}, and MMBench-CN/EN~\cite{liu2024mmbench}, covering a diverse range of perception and reasoning tasks. MMVet targets compositional and multi-step reasoning through open-ended questions. MME assesses both perceptual and cognitive abilities via structured tasks. MMStar emphasizes fine-grained perception and complex reasoning. MMBench-CN and MMBench-EN are standardized multiple-choice benchmarks in Chinese and English, respectively, designed for systematic evaluation of perception and reasoning capabilities. 
We report the official metrics provided by each benchmark. Additional details are provided in Appendix \S\ref{appendix:datasets}.

\noindent \textbf{Baselines.}
We evaluate our method on four state-of-the-art MLLMs, including Qwen2.5VL-3B, Qwen2.5VL-7B~\cite{Bai2025Qwen25VLTR}, LLaVA-NeXT-7B, and LLaVA-NeXT-13B~\cite{liu2024llavanext}. We compare our approach with two test-time training methods, MM-UPT~\cite{wei2025first} and TLM~\cite{hu2025test}. Originally designed for LLMs, TLM performs test-time training by directly minimizing input perplexity without requiring external supervision. When extended to multimodal settings, TLM updates model parameters by optimizing the likelihood of the test-time query $q$. More details are provided in Appendix \S\ref{appendix:baseline}.

\noindent \textbf{Implementation Details.} 
We adopt  Low-Rank Adaptation (LoRA)~\cite{hu2022lora} to update the model parameters during test-time training. Specifically, we set the LoRA rank to 8, the scaling factor to 16, and the dropout rate to 0.1. LoRA parameters are applied to all layers of the model.
We use the AdamW~\cite{loshchilov2017decoupled} optimizer with a learning rate selected from $\{1\times10^{-6}, 1\times10^{-5}, 1\times10^{-4}\}$. During test-time training, the model is updated for 8 gradient steps for each test instance.
Additional implementation details are provided in Appendix \S\ref{appendix:add_imp_detail}.

\subsection{Main Results}\label{sec:main_result}
As shown in Table~\ref{table:main_result}, our method \textbf{consistently improves} the performance of \textbf{all} baseline models across all benchmarks, demonstrating its broad applicability. For example, when applied to LLaVA-NeXT-7B, our method improves MMVet from 42.60 to 44.60, MMStar from 37.53 to 38.00, and MME from 1788 to 1811. Similar trends are observed on Qwen2.5VL-3B, where our method improves MMVet from 61.80 to 62.50, MMStar from 55.06 to 55.86, and MME from 2191 to 2229. Crucially, on more powerful scale architectures like Qwen2.5VL-7B and LLaVA-NeXT-13B, our method robustly maintains consistent performance gains across all benchmarks.

Compared to existing TTT approaches, our method achieves stronger and more stable improvements. In particular, MM-UPT often yields limited gains or even causes performance degradation on several benchmarks. For example, on Qwen2.5VL-3B, MM-UPT reduces MMStar from 55.06 to 54.67, whereas our method improves it to 55.86. Similarly, TLM exhibits inconsistent behavior across benchmarks. On Qwen2.5VL-7B, TLM achieves no improvement on MMStar and even degrades performance on MME from 2327 to 2322. In sharp contrast, our method consistently achieves improvements across all benchmarks.

Overall, the results show that effective and generalizable performance gains can be achieved through a simple yet robust test-time training strategy.

\subsection{Observations}\label{sec:obs}
\begin{figure}[t!]
    \centering
    \includegraphics[width=0.48\textwidth,
    trim=1.1cm 0.0cm 0.1cm 0cm, clip]{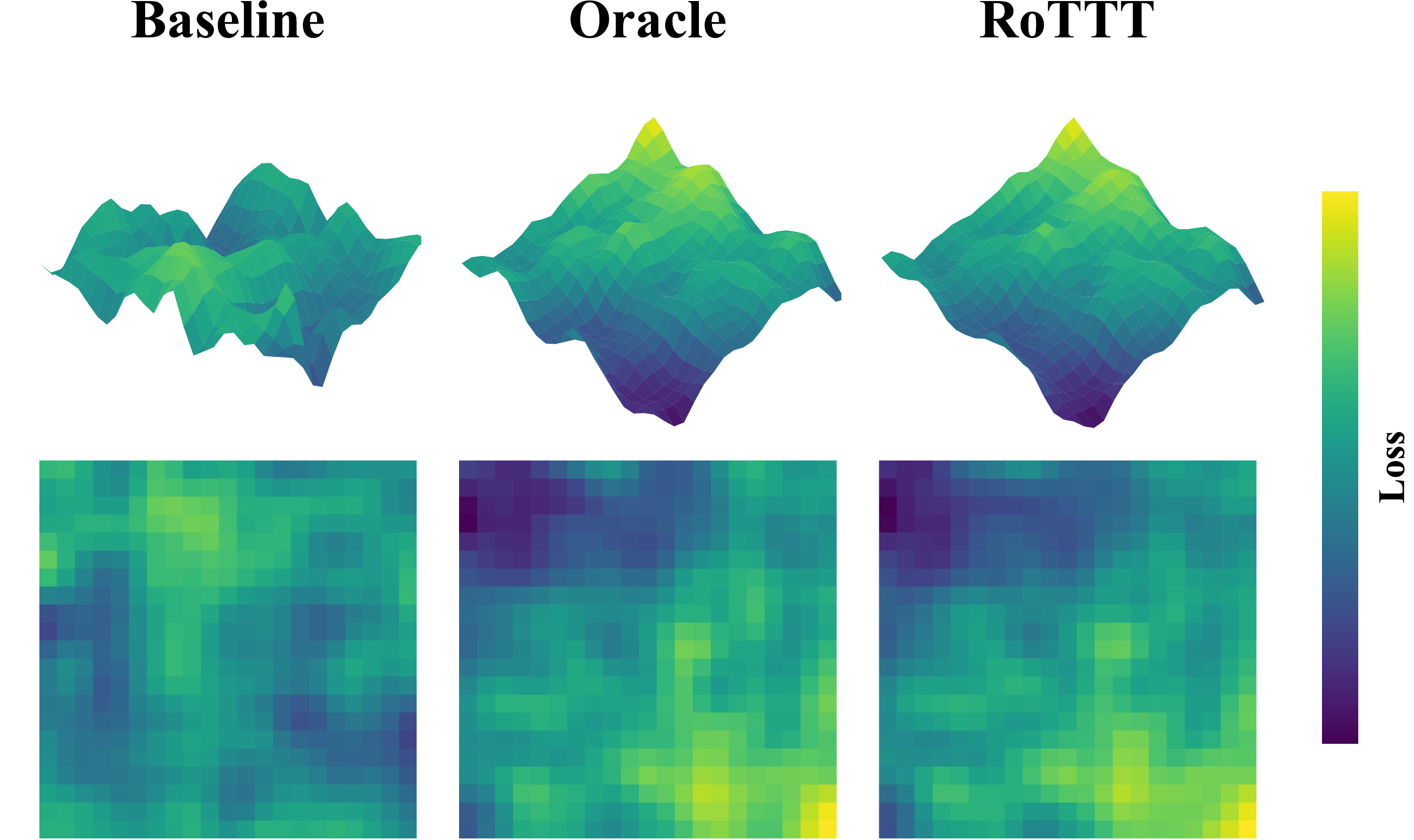}
    \vspace{-10pt}
    \caption{\textbf{Loss Landscape Visualization} of Qwen2.5-VL-3B on an MMVet sample. The top row shows the loss landscapes, while the bottom row presents the corresponding 2D heatmaps. From left to right, the columns correspond to the baseline model, the oracle model, and the RoTTT-adapted model.}
    \vspace{-10pt}
    \label{fig:loss_landscape}
\end{figure}

\noindent \textbf{Loss Landscape Analysis.} To investigate the optimization dynamics of RoTTT, we visualize and compare the loss landscapes~\cite{li2018visualizing,ma2022rethinking,liu2025re} across the baseline model without adaptation, an oracle model fine-tuned on the ground-truth label, and the model adapted via RoTTT. The oracle model serves as a reference for the ideal optimization direction guided by true supervision.
As illustrated in Fig.~\ref{fig:loss_landscape}, the loss landscape of RoTTT closely resembles that of the oracle model, while both differ significantly from the baseline.
This observation suggests that, despite not using ground-truth labels, RoTTT induces optimization trajectories that align well with those guided by true supervision. Consequently, rotation-based self-supervision provides a reliable surrogate signal that approximates the oracle optimization direction, leading to more stable updates and improved performance during test-time training. More details are provided in Appendix \S\ref{appendix:loss_landscape}.

\begin{figure}[t!]
    \centering
    \vspace{-12pt}
    \includegraphics[width=0.48\textwidth]{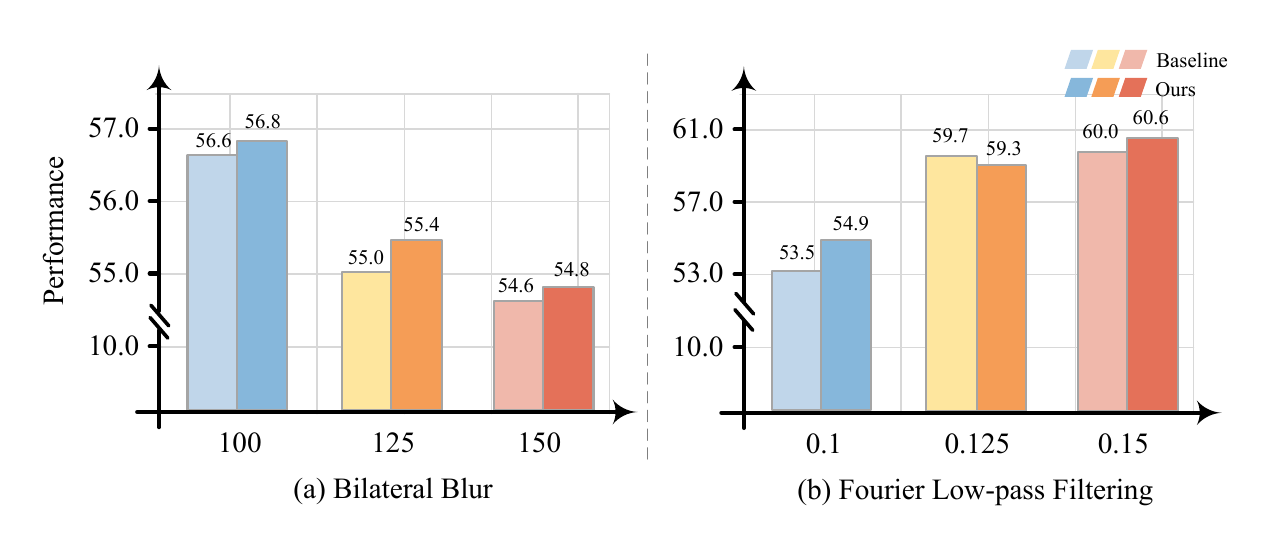}
    \vspace{-20pt}
    \caption{\textbf{Blur perturbation analysis.} RoTTT exhibits consistently smaller performance drops than the baseline under various blur perturbations}
    \vspace{-10pt}
    \label{fig:blur_analysis}
\end{figure}

\noindent \textbf{Blur Perturbation Analysis.} We further investigate whether RoTTT learns spatially-aware representations through a blur perturbation analysis using Qwen2.5VL-3B on MMVet. Blur perturbations primarily suppress high-frequency local texture details while largely preserving global spatial structure and object layout. Therefore, robustness under blur perturbation can serve as an indicator of whether a model relies more on local textures or on spatial and structural cues. Specifically, after performing RoTTT adaptation, we apply blur perturbations to the test images during inference and compare the performance degradation of the baseline and RoTTT-adapted models. We consider two types of perturbations, including bilateral blur~\cite{tomasi1998bilateral} and Fourier low-pass filtering~\cite{nussbaumer1981fast,almeida2002fractional}, each evaluated under three different intensities.
As shown in Fig.~\ref{fig:blur_analysis}, although blur degrades the performance of all models, RoTTT consistently exhibits smaller performance drops in five out of six settings. For example, under bilateral blur with a strength of 125, the baseline drops to 55.0, while RoTTT maintains 55.4. Similar trends are observed under Fourier low-pass perturbations, where RoTTT achieves consistently superior robustness under most perturbation levels.
These results suggest that RoTTT encourages the model to rely more on spatial and structural cues rather than local textures. We attribute this to the rotation prediction objective, which inherently requires understanding object orientation and spatial configuration, thereby promoting more spatially-aware representations during test-time training.

\begin{table}[b]
\centering
\vspace{-10pt}
\centering
\resizebox{0.45\textwidth}{!}{
\begin{tabular}{l||rr|rr}
\hline \thickhline
\rowcolor{mygray}
& \multicolumn{2}{c|}{Qwen2.5VL-3B} & \multicolumn{2}{c}{LLaVA-NeXT-7B} \\
\rowcolor{mygray}
Label&MMVet &MME &MMVet &MME\\
\hline\hline
simplified &62.40  &2214 &44.40 &1788   \\
structured &\textbf{62.50}  &\textbf{2229} &\textbf{44.60} &\textbf{1811}   \\
\hline
\end{tabular}
}
\caption{\textbf{Impact of Prediction Label}. Structured labels consistently outperform simplified labels.}
\label{table:pre_label}
\end{table}

\subsection{Ablation Study}\label{sec:ab_study}
We conduct an ablation study on Qwen2.5VL-3B and LLaVA-NeXT-7B using MMvet and MME.

\noindent\textbf{Impact of Prediction Label.} 
We examine the impact of label design for the rotation prediction task. Specifically, we compare a structured formulation, $\{\textit{direction}, \textit{angle\_range}, \textit{angle}\}$, with a simplified variant using only $\{\textit{angle}\}$. Both settings are optimized using the same autoregressive objective, ensuring a fair comparison. As shown in Table~\ref{table:pre_label}, the structured formulation consistently achieves superior performance across all benchmarks.

The performance gap stems from the richer, multi-level supervision provided by the structured labels, simplifying the learning problem into manageable sub-tasks. This coarse-to-fine decomposition facilitates more efficient optimization during the limited update steps of TTT. In contrast, predicting only the angle lacks sufficient auxiliary signals, making the learning process more challenging and less stable. These findings highlight the value of structured prediction targets in TTT.

\noindent\textbf{Impact of LoRA Rank.} To investigate the impact of LoRA rank, we evaluate performance across different rank settings (\ie, 4, 8, and 16). As shown in Table~\ref{table:lora_rank}, we observe a performance peak at rank 8, which yields optimal results for both Qwen2.5VL-3B and LLaVA-NeXT-7B across the MMVet and MME benchmarks. Both smaller and larger ranks (\eg, 4 and 16) result in degraded performance, suggesting that insufficient or excessive adaptation capacity can hinder test-time optimization. A low rank limits the model’s ability to capture instance adjustments, while a high rank may introduce optimization instability under limited update steps. Consequently, we adopt a rank of 8 as the default configuration for all other experiments.

\begin{table}[t]
\centering
\centering
\resizebox{0.45\textwidth}{!}{
\begin{tabular}{l||rr|rr}
\hline \thickhline
\rowcolor{mygray}
LoRA& \multicolumn{2}{c|}{Qwen2.5VL-3B} & \multicolumn{2}{c}{LLaVA-NeXT-7B} \\
\rowcolor{mygray}
Rank&MMVet &MME &MMVet &MME\\
\hline\hline
4 &62.00  &2217 &43.40 &1788   \\
8 &\textbf{62.50}  &\textbf{2229} &\textbf{44.60} &\textbf{1811}   \\
16 &62.30  &2216 &44.30 &1802  \\
\hline
\end{tabular}
}
\caption{\textbf{Impact of LoRA Rank}.The results show that rank 8 consistently achieves the best performance across models and benchmarks, while both lower and higher ranks lead to suboptimal results.}
\label{table:lora_rank}
\end{table}

\noindent \textbf{Impact of LoRA Position.} We then study the effect of applying LoRA to different components of the MLLM architecture, including the vision encoder, the LLM backbone, the last layers, and all layers. As shown in Table~\ref{table:lora_palce}, applying LoRA to all layers achieves the best performance for both Qwen2.5VL-3B and LLaVA-NeXT-7B on MMVet and MME. This suggests that effective test-time training necessitates coordinated and holistic updates across both visual and linguistic modules. In contrast, restricting LoRA to isolated components (\ie, the vision encoder, the LLM backbone, or the last layers) provides insufficient adaptation capacity,
limiting the model’s ability to jointly refine visual representations and multimodal reasoning. This observation underscores the importance of global LoRA updates during TTT. Therefore, we apply LoRA to all layers in our experiments.

\begin{table}[t]
\centering
\centering
\resizebox{0.45\textwidth}{!}{
\begin{tabular}{l||rr|rr}
\hline \thickhline
\rowcolor{mygray}
LoRA& \multicolumn{2}{c|}{Qwen2.5VL-3B} & \multicolumn{2}{c}{LLaVA-NeXT-7B} \\
\rowcolor{mygray}
Position&MMVet &MME &MMVet &MME\\
\hline\hline
Vision &62.30  &2201 &42.80 &1789   \\
LLM &62.20  &2206 &44.00 &1788   \\
Last Layer &61.80  &2217 &43.80 &1788  \\
All &\textbf{62.50}  &\textbf{2229} &\textbf{44.60} &\textbf{1811}  \\
\hline
\end{tabular}
}
\caption{\textbf{Impact of LoRA Position}. We evaluate the effect of applying LoRA to different components of the MLLM.
The results show that applying LoRA to all layers consistently achieves the best performance across models and benchmarks.}
\vspace{-10pt}
\label{table:lora_palce}
\end{table}

\noindent \textbf{Computational Efficiency.} We compare the computational cost of our approach with GRPO-based MM-UPT. As shown in Table~\ref{table:comp_cost}, our method consistently demonstrates superior computational efficiency across all benchmarks and models. Specifically, on Qwen2.5VL-3B, our method reduces computational cost by 3.2$\times$ to 6$\times$. 
The margin becomes significantly larger on LLaVA-NeXT-7B, achieving up to a 
30$\times$ speedup, which highlights a substantial efficiency advantage. 

This difference is primarily attributed to the underlying optimization strategy: while GRPO-based MM-UPT relies on multiple sampling, our method performs only a single forward and backward pass per update. In addition, the computational gap becomes more pronounced as the sampling size increases and as models demand more GPU memory.
Overall, these results demonstrate that our method not only improves performance but also offers substantial computational savings, making it more scalable and practical for TTT.
Additional results are provided in Appendix \S\ref{appendix:add_comp_cost_result}.

\begin{table}[t]
\centering
\centering
\resizebox{0.48\textwidth}{!}{
\setlength{\tabcolsep}{2pt}
\begin{tabular}{l||rrr|rrr}
\hline \thickhline
\rowcolor{mygray}
& \multicolumn{3}{c|}{Qwen2.5VL-3B} & \multicolumn{3}{c}{LLaVA-NeXT-7B} \\
\rowcolor{mygray}
Dataset &Ours &MM-UPT & $\times$ Cost&Ours &MM-UPT&$\times$ Cost\\
\hline\hline
MMStar &18.1  &61.2  &${\sim}3.4\times$ &24.2 &701.7 &${\sim}29.0\times$    \\
MME &28.9  &174.7  & ${\sim}6.0\times$ &40.5 &1214.6  & ${\sim}30.0\times$  \\
MMbench-CN &22.8  &72.8  & ${\sim}3.2\times$ &29.2 &752.8 & ${\sim}25.8\times$    \\
MMbench-EN &21.3  &72.1  &${\sim}3.4\times$ &30.0 &812.0  & ${\sim}27.1\times$  \\
\hline
\end{tabular}
}
\caption{\textbf{Comparison of Computational Efficiency}. We report the total test-time training cost in GPU hours for each benchmark. The ``$\times$ Cost'' column indicates the relative speedup factor. RoTTT consistently reduces computational overhead across all benchmarks and models, offering a more resource-efficient paradigm for on-the-fly MLLM adaptation.}
\vspace{-10pt}
\label{table:comp_cost}
\end{table}

%% file: sec/6_conclusion.tex
\section{Conclusion}
In this paper, we present RoTTT (\textbf{Ro}tation-based \textbf{T}est-\textbf{T}ime \textbf{T}raining), a simple yet effective test-time training approach for MLLMs that leverages rotation-based self-supervision for instance-specific adaptation. It has several key merits: \textbf{i)} introducing a lightweight and reliable test-time objective without requiring additional annotations or multi-sample optimization; \textbf{ii)} improving robustness under distribution shift through spatially-aware representation refinement; and \textbf{iii)} demonstrating consistent gains across diverse benchmarks and model architectures with high computational efficiency.
Overall, our results suggest that transformation-based self-supervision offers a practical and promising direction for efficient and robust test-time training in MLLMs.

\label{sec:conclusion}

%% file: sec/limitation.tex
\section*{Limitations}
While our approach provides an efficient and label-free paradigm for test-time training, it still introduces an inherent online optimization overhead during inference, a characteristic common to all TTT frameworks. One possible mitigation strategy is to employ selective adaptation, where test-time updates are triggered only when the model encounters low-confidence inputs or severe domain shifts. Such a mechanism could substantially reduce the amortized inference overhead while preserving the benefits of TTT.

\section*{Ethical Considerations}
We conform to the ACL Code of Ethics and further show the consent to our work below. All datasets used in this study are publicly available and released under permissive licenses, and all the models are publicly available (see Appendix~\S\ref{appendix:License} for Asset License and Consent). We would like to state that the contents in the dataset do NOT represent our views or opinions, and our paper does not involve crowdsourcing or research with human subjects. 

%% file: sec/acknowledge.tex
\section*{Acknowledgments}
This work was funded by the National Science Foundation under Grant No. 2450068. The computing in this work was partially conducted through the NCSA Delta CPU and GPU through allocation CIS250460 from the Advanced Cyberinfrastructure Coordination Ecosystem: Services \& Support (ACCESS) program, which is supported by U.S. National Science Foundation grants No. 2138259, No. 2138286, No. 2138307, No. 2137603, and No. 2138296.

%% file: sec/X_suppl.tex
\clearpage

\appendix
\renewcommand{\thesection}{S\arabic{section}}
\renewcommand{\thetable}{S\arabic{table}}
\renewcommand{\thefigure}{S\arabic{figure}}
\setcounter{table}{0}
\setcounter{figure}{0}
\setcounter{section}{0}
\centerline{\textbf{SUMMARY OF THE APPENDIX}} 
\vspace{0.5em}

This appendix contains additional experimental results and discussions of our work, organized as:
\begin{itemize}
\setlength{\itemsep}{0pt}
  \item \S\ref{appendix:more_rl} provides \textbf{more related works} on TTT.
  \item \S\ref{appendix:datasets} details the \textbf{datasets} utilized in the study.
  \item \S\ref{appendix:baseline} details the \textbf{baseline models} utilized in the study.
  \item \S\ref{appendix:add_imp_detail} provides additional \textbf{implementation details.} 
  \item \S\ref{appendix:vis_result} provides additional \textbf{results visualizations}.
  \item \S\ref{appendix:mme_result} provides \textbf{detailed MME results}.
  \item \S\ref{appendix:loss_landscape} provides additional \textbf{loss landscape analysis.} 
  \item \S\ref{appendix:add_comp_cost_result} provides additional \textbf{computational cost results.} 
  \item \S\ref{appendix:sec:reproduce} includes the \textbf{reproducibility statement and pseudo code} of our method. 
  \item \S\ref{appendix:sec:technical_contributions} highlights the \textbf{technical contributions} of our method.
  \item \S\ref{appendix:License} offers a \textbf{summary of licenses and consent}, and lists usage terms for all models and datasets. 
  
  \item \S\ref{appendix:ethics} includes additional discussions on \textbf{ethics concerns}.
  \item \S\ref{appendix:Discussion} discusses on \textbf{future directions}, highlighting potential areas for further research.
  \item \S\ref{appendix:disclosure} provides an \textbf{AI disclosure}, and notes that AI assistance was limited to grammar checking. 
\end{itemize}

\section{More Related Works}
\label{appendix:more_rl}
\subsection{More Discussions on Test-time Training}
Recent work has explored test-time training for vision-language models (VLM) such as CLIP, aiming to improve robustness under distribution shift~\cite{dong2025adapting}. These approaches can be broadly categorized into prompt adaptation, entropy minimization, and distribution alignment methods. Prompt adaptation methods~\cite{shu2022test, yoonc, zhu2024efficient, zhang2024historical, feng2023diverse, ma2023swapprompt, zanella2024test, zhang2024robust, wang2025tapt, han20232vpt, wang2024m2pt} update textual or visual prompts at inference time, providing a lightweight alternative to tuning model parameters. Entropy minimization approaches~\cite{sheng2025r, sharifdeen2025tpt, a2025clipartt, imam2025test, wu2026adaptive, osowiechi2024watt} adapt models by encouraging confident predictions on test samples without requiring ground-truth labels~\cite{wang2020tent}. Distribution alignment methods~\cite{lafon2026cliptta, jung2023cafa, abdul2023align, tong2025test, maharana2025batclip, zhou2023test} further reduce domain gaps by aligning feature distributions between source and target domains. 

However, such approaches are primarily developed for CLIP-style models, which rely on fixed feature extractors and relatively simple prediction heads. In contrast, Multimodal Large Language Models (MLLMs) adopt an autoregressive generative paradigm, involving complex cross-modal reasoning and token-level prediction. This fundamental difference makes it challenging to directly apply existing adaptation strategies.

\section{Datasets}
\label{appendix:datasets}
In Table~\ref{table:app_datasets}, we provide statistical information (\ie test sample number) of five multimodal understanding benchmarks: MMVet~\cite{yu2024mm}, MME~\cite{fu2023mme}, MMStar~\cite{chen2024we}, and MMBench-CN/EN~\cite{liu2024mmbench}.
\begin{table}[ht]
\vspace{-2mm}
\begin{adjustbox}{width=0.99\width,center}
\begin{tabular}{l||c}
\hline \thickhline
\rowcolor{mygray}
Dataset &Test Number\\
\hline \hline

MMVet   & 218\\
MME  &2,373\\
MMStar   &1,500\\
MMBench-CN   &1,784\\
MMBench-EN  &1,784\\
\hline
\end{tabular}
\end{adjustbox}
\caption{\textbf{Benchmarks Statistical Details.}}
\label{table:app_datasets}
\end{table}

For MMVet, evaluation is conducted using GPT-4, and the scores are reported as percentages with a maximum of 100\%.
For MME, results are reported as the sum of perception and cognition scores, corresponding to the total number of correct answers.
For MMStar, MMBench-CN, and MMBench-CN, performance is measured using accuracy, reported as a percentage with a maximum score of 100\%.
Notably, MMBench-CN and MMBench-CN adopt Circular Evaluation~\cite{liu2024mmbench} to evaluate results. Circular Evaluation improves the robustness of multiple-choice evaluation by circularly shifting answer choices across multiple inference passes. A prediction is considered correct only if the model answers all shifted versions correctly, reducing random-guessing effects and positional bias in evaluation.

\begin{figure*}[t]
    \centering
    \vspace{-6pt}
    \includegraphics[width=\textwidth]{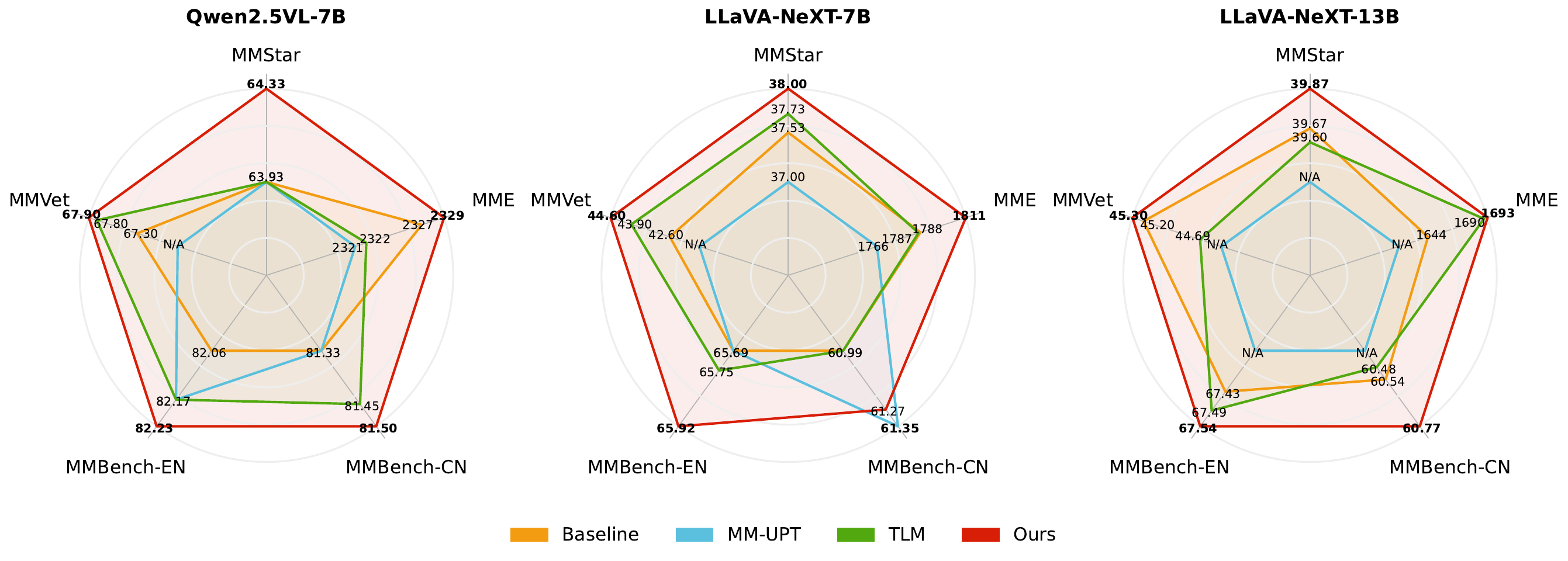}
    \vspace{-6pt}
    \caption{\textbf{Additional results visualizations} for Qwen2.5VL-7B, LLaVA-NeXT-7B, and LLaVA-NeXT-13B.}
    \vspace{-10pt}
    \label{fig:radar_three}
\end{figure*}

\section{Baseline Models}
\label{appendix:baseline}
We evaluate our method on four state-of-the-art MLLMs, which span diverse architectural designs and scales:
\begin{itemize}[noitemsep, left=1pt]
\item \textbf{Qwen2.5-VL-3B}~\cite{Bai2025Qwen25VLTR} is a 3B-scale multimodal large language model built upon the Qwen2.5~\cite{qwen2025qwen25technicalreport} LLM backbone, designed for efficient vision-language understanding in resource-constrained settings. 
\item \textbf{Qwen2.5-VL-7B}~\cite{Bai2025Qwen25VLTR} is a larger 7B-scale variant of Qwen2.5-VL.
\item \textbf{LLaVA-NeXT-7B}~\cite{liu2024llavanext} is a 7B-scale multimodal large language model built upon the Mistral-7B~\cite{jiang2023mistral7b} backbone.
\item \textbf{LLaVA-NeXT-13B}~\cite{liu2024llavanext} is a larger 13B-scale variant built upon the Vicuna-13B~\cite{vicuna2023} backbone.
\end{itemize}

We compare our approach with two test-time training methods, MM-UPT~\cite{wei2025first} and TLM~\cite{hu2025test}.
Notably, TLM was originally developed for LLMs, and we extend it to the multimodal setting for fair comparison with MLLMs. In the original LLM setting, TLM performs test-time training by minimizing input perplexity without requiring external supervision. In the multimodal setting, we implement TLM by updating model parameters through minimizing the perplexity of the input query (\ie text part) alone.
\section{Additional Implementation Details.}
\label{appendix:add_imp_detail}
MM-UPT use GRPO-style to update the model. The group size is 8. A larger group size will lead to GPU OOM. For Qwen2.5-VL-3B and Qwen2.5-VL-7B, we train on one GPU. However, due to the higher memory consumption of LLaVA-NeXT-7B, four GPUs are required for stable training. For LLaVA-NeXT-13B, MM-UPT consistently encounters OOM issues even under multi-GPU settings. Therefore, we do not report MM-UPT results on LLaVA-NeXT-13B. 

We additionally observe that baseline LLaVA-NeXT-13B performs worse than baseline LLaVA-NeXT-7B on several benchmarks, including MMVet and MMBench-CN. We attribute this behavior to differences in the underlying LLM backbones rather than model scale alone. Specifically, LLaVA-NeXT-7B is built upon Mistral-7B, whereas LLaVA-NeXT-13B uses Vicuna-13B. A similar trend is also reported in ~\cite{li2024llavanext-strong}, where the LLaVA-NeXT-13B model underperforms the LLaVA-NeXT-7B (Mistral-7B) variant on the MME cognition tasks, further supporting our observation.

The rotation prediction prompt $p_{\text{rot}}$ used in our method is:
``The second image is a rotated version of the first image. Answer in three parts: 
(1) Direction: choose one from `clockwise' or `counter-clockwise'; 
(2) Angle range: choose one from `small' (1--15$^\circ$), `medium' (16--30$^\circ$), or `large' (31--45$^\circ$); 
(3) Angle: the exact absolute rotation angle in degrees (0--45). 
Output only a JSON object: \{"direction": "<clockwise|counter-clockwise>", ``angle\_range": "<small|medium|large>", ``angle": <int>\}.''

\section{Additional Results Visualizations.}
\label{appendix:vis_result}
We provide additional visualizations comparing RoTTT with MM-UPT and TLM in Fig.~\ref{fig:radar_three} . The results consistently show that RoTTT achieves stronger and more stable improvements across benchmarks.

\section{Detailed MME Results.}
\label{appendix:mme_result}
MME evaluates both perception and cognition capabilities of MLLMs. We report the perception and cognition scores separately, along with the overall performance in Table~\ref{table:mme_result}.

\begin{table}[H]
\begin{adjustbox}{width=0.79\width,center}
\begin{tabular}{l||ccc} 
\hline \thickhline
\rowcolor{mygray}
Method &  Perception & Cognition &  Overall   \\
\hline \hline

  $\text{Qwen2.5-VL-3B}$ \textcolor{lightgray}{\scriptsize{[arXiv25]}}   &1588  &603  &2191  \\
 + MM-UPT \textcolor{lightgray}{\scriptsize{[NeurIPS25]}} &1601  &617  &2218  \\
 + TLM \textcolor{lightgray}{\scriptsize{[ICML25]}} &1601  &622  &2223  \\
  \rowcolor{mygray}
 + Ours&1608  &621  &2229  \\
 \hline
  $\text{Qwen2.5-VL-7B}$ \textcolor{lightgray}{\scriptsize{[arXiv25]}}   &1715  &612  &2327  \\
 + MM-UPT \textcolor{lightgray}{\scriptsize{[NeurIPS25]}} &1709  &612  &2321  \\
 + TLM \textcolor{lightgray}{\scriptsize{[ICML25]}} &1710  &612  &2322\\
  \rowcolor{mygray}
 + Ours&1717  &612  &2329  \\
 \hline
  $\text{LLaVA-NeXT-7B}$ \textcolor{lightgray}{\scriptsize{[arXiv24]}}   &1492  &296  &1788  \\
 + MM-UPT \textcolor{lightgray}{\scriptsize{[NeurIPS25]}} &1487  &279  &1766  \\
 + TLM \textcolor{lightgray}{\scriptsize{[ICML25]}} &1489  &298  &1787  \\
  \rowcolor{mygray}
 + Ours&1508  &303  &1811  \\
 \hline
  $\text{LLaVA-NeXT-13B}$ \textcolor{lightgray}{\scriptsize{[arXiv24]}}   &1392  &272  &1664  \\
 + MM-UPT \textcolor{lightgray}{\scriptsize{[NeurIPS25]}} &-  &-  &- \\
 + TLM \textcolor{lightgray}{\scriptsize{[ICML25]}} &1403  &287  &1690  \\
  \rowcolor{mygray}
 + Ours&1399  &294  &1693  \\
 \hline

\end{tabular}
\end{adjustbox}

\caption{\textbf{MME Benchmark Performance.} We report perception, cognition, and overall scores on the MME benchmark.}
\vspace{-3mm}
\label{table:mme_result}
\end{table}

\section{Additional Loss Landscape Analysis.}
\label{appendix:loss_landscape}
\begin{figure}[H]
    \centering
    \vspace{-6pt}
    \includegraphics[width=0.48\textwidth,
    trim=1.1cm 0.0cm 0.1cm 0cm, clip]{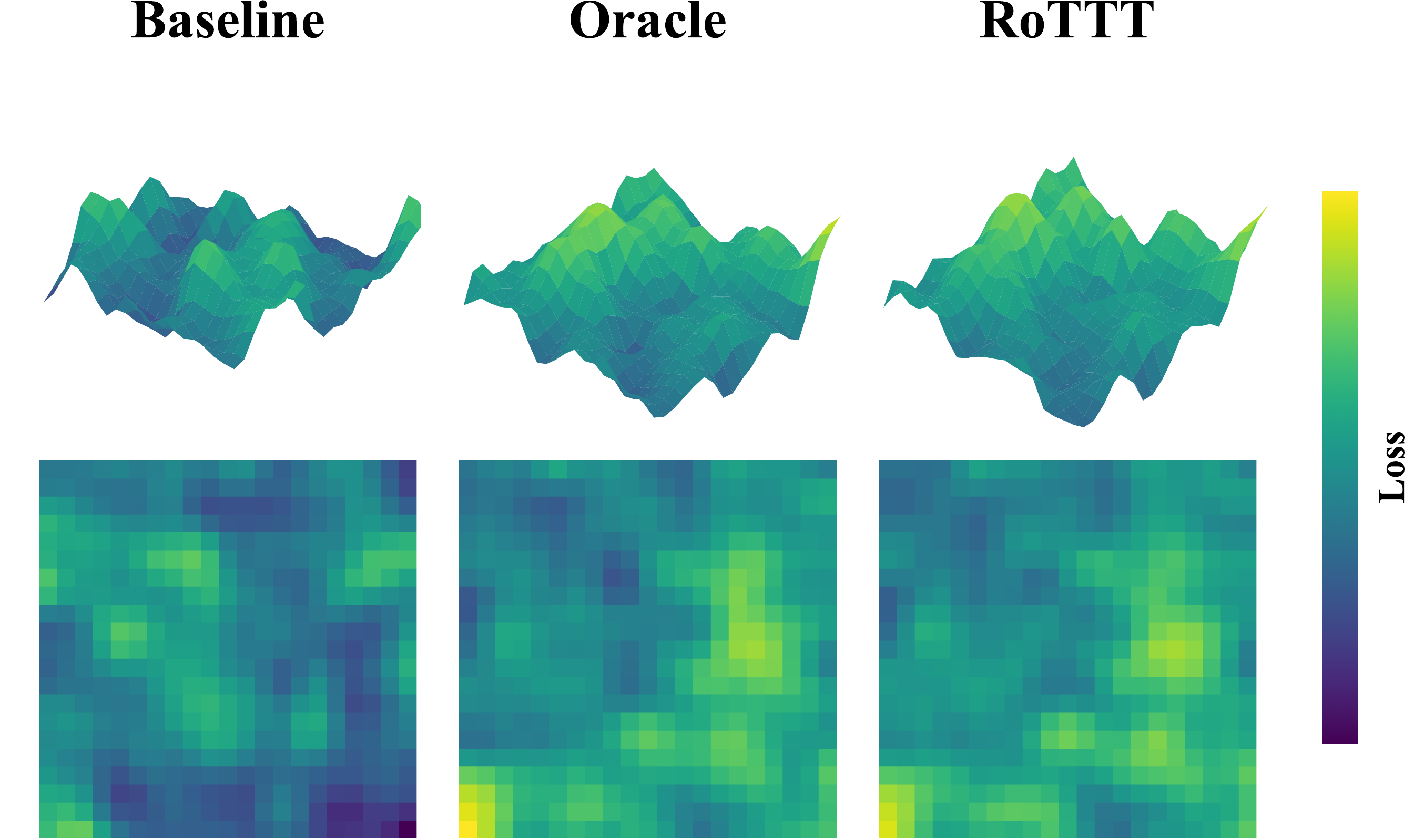}
    \vspace{-6pt}
    \caption{\textbf{Additional Loss Landscape Visualization.}}
    \vspace{-5pt}
    \label{fig:loss_landscape_2}
\end{figure}

To analyze the optimization behavior of RoTTT, we visualize the loss landscape using a standard interpolation-based approach~\cite{li2018visualizing}. Starting from the model parameters, we evaluate the loss over a grid of points generated by perturbing the parameters along two sampled directions. The loss is computed for different settings, including the baseline model, the oracle model, and the RoTTT-adapted model. The resulting loss values are then plotted to obtain a loss landscape, enabling qualitative comparison of optimization behavior across different strategies. We additionally plot the loss landscape for another MMVet test sample on the Qwen2.5-VL-3B model, as shown in Fig.~\ref{fig:loss_landscape_2}. Notably, the loss landscape of RoTTT aligns closely with that of the oracle model, whereas both deviate significantly from the baseline.

\section{Additional Computational Cost Results.}
\label{appendix:add_comp_cost_result}
We provide additional computational cost comparisons on Qwen2.5-VL-7B to further evaluate the efficiency of our method. As shown in Table~\ref{table:comp_cost_2}, RoTTT requires significantly less computation than MM-UPT, which is consistent with our observation in \S\ref{sec:ab_study}

\begin{table}[t]
\centering
\centering
\resizebox{0.45\textwidth}{!}{
\begin{tabular}{l||rrr}
\hline \thickhline
\rowcolor{mygray}
& \multicolumn{3}{c}{Qwen2.5-VL-7B} \\
\rowcolor{mygray}
Dataset &Ours &MM-UPT & $\times$ Cost\\
\hline\hline
MMStar &16.9  &72.0  &${\sim}4.3\times$   \\
MME &29.3  &178.1  & ${\sim}6.1\times$ \\
MMbench-CN &21.9  &85.0  & ${\sim}3.9\times$  \\
MMbench-EN &22.2  &89.6  &${\sim}4.0\times$ \\
\hline
\end{tabular}
}
\vspace{-6pt}
\caption{\textbf{Comparison of Computational Efficiency} for Qwen2.5-VL-7B.}
\label{table:comp_cost_2}
\end{table}

\section{Reproducibility}
\label{appendix:sec:reproduce}
RoTTT is implemented in Pytorch~\cite{NEURIPS2019_9015}. Experiments are conducted on NVIDIA RTX 6000 Ada GPUs. 
To guarantee reproducibility, our full implementation shall be publicly released upon paper acceptance. 
We provide the pseudo code of RoTTT in Algorithm~\ref{alg:tt_rotation}.

\section{Technical Contributions}
\label{appendix:sec:technical_contributions}

Our study presents three principal technical contributions: 
\begin{itemize}
    \item We introduce RoTTT, a simple and efficient test-time training framework for MLLMs that enables instance-specific adaptation during inference without requiring external supervision or costly multi-sample optimization.
    \item We propose a rotation-based self-supervised objective that leverages the rotation angle as a reliable supervision signal. By formulating rotation prediction as a structured textual reasoning task, our method promotes spatial-aware representation learning while remaining fully compatible with the native autoregressive objective of MLLMs.
    
    \item Extensive experiments across multiple multimodal understanding benchmarks and MLLM architectures demonstrate that RoTTT consistently improves performance while significantly reducing computational overhead compared to existing MLLM TTT approaches.

\end{itemize}

\section{Asset License and Consent}
\label{appendix:License}
All models and datasets used in this work are publicly available.  
We strictly comply with their original licenses and use them only for non-commercial academic research.  
The contents of datasets do not represent our views or opinions.

\noindent\textbf{Models.}
We utilize four open-source models:
Qwen2.5-VL-3B (Qwen RESEARCH LICENSE AGREEMENT),
Qwen2.5-VL-7B (Qwen RESEARCH LICENSE AGREEMENT),
LLaVA-NeXT-7B (LLAMA 2 Community License),
LLaVA-NeXT-13B (LLAMA 2 Community License).
All licenses permit academic research use; detailed terms are available via the original model repositories.  

\noindent\textbf{Datasets.}  
MMVet is licensed under Apache License 2.0, 
MMStar is licensed under Apache License 2.0, 
and MMBench-CN/EN is licensed under Apache License 2.0. Notably, MME does not provide a publicly specified license, but it is widely used for academic research purposes.

\noindent\textbf{Consent.}  
Our study does not involve crowdsourcing or human subjects. All results are derived from publicly available models and datasets.  

\section{Ethics Concerns}
\label{appendix:ethics}
While RoTTT improves robustness and enables on-the-fly adaptation for MLLMs, it also introduces a potential risk: test-time training updates model parameters during inference, which may lead to unintended behavior if the adaptation process overfits to noisy or adversarial inputs \cite{li2026evolving}.

\section{Future Direction}\label{appendix:Discussion}
A promising direction for future work is to extend RoTTT to broader multimodal settings, such as video understanding \cite{lin2023video, wang2025test} and embodied agents \cite{liu2026fly, Assran2025V-JEPAPlanning, lu2025vla, wang2025vla, guo2026priorvla}. In these domains, inputs are inherently dynamic and temporally structured, requiring models to adapt continuously to evolving environments. Extending our rotation-based test-time training to these settings could enable robust, flexible adaptation, thereby enhancing performance in complex, real-world tasks.

Another important direction is to improve the safety and stability of test-time training against adversarial noise. To mitigate the risk of unintended behaviors caused by overfitting to noisy or adversarial inputs, incorporating robust regularization or lightweight safety alignment strategies during test-time training would be valuable extensions.

\section{AI Disclosure}
\label{appendix:disclosure}
We acknowledge the use of GPT-5.5 for grammar checking only. The model was employed to correct grammatical errors while ensuring the original meaning and intent of the text remained unchanged.